# Agentic LLM Workflow for MR Spectroscopy Volume-of-Interest Placements in Brain Tumors

Sangyoon Lee[1], Francesca Branzoli[2], Małgorzata Marjańska[1], Patrick Bolan[1]

[1] Center for Magnetic Resonance Research, University of Minnesota
[2] ICM – Paris Brain Institute

**Abstract.** Magnetic resonance spectroscopy (MRS) provides clinically valuable metabolic characterization of brain tumors, but its utility depends on accurate placement of the spectroscopy volume-of-interest (VOI). However, VOI placement typically has a broad operating window: for a given tumor there are multiple possible VOIs that would lead to high-quality MRS measurements. Thus, a VOI placement can be tuned for clinician preference, case-specific anatomy, and clinical priorities, which leads to high inter-operator variability, especially for heterogeneous tumors. We propose an agentic large language model (LLM) workflow that decomposes VOI placement into generation of diverse candidate VOIs, from which the LLM selects an optimal one based on quantitative metrics. Candidate VOIs are generated by vision transformer-based placement models trained with different objective function preferences, which allows selection from acceptable alternatives rather than a single deterministic placement. On 110 clinical brain tumor cases, the agentic workflow achieves improved solid tumor coverage and necrosis avoidance depending on the user preferences compared to the general-purpose expert placements. Overall, the proposed workflow provides a strategy to adapt VOI placement to different clinical objectives without retraining task-specific models.

## 1 Introduction

Magnetic resonance spectroscopy (MRS) provides noninvasive metabolic characterization of brain tissue and can be used for diagnosis, treatment planning, and longitudinal assessment [1]. For example, MRS can help differentiate clinically relevant subregions such as solid tumor, infiltrative tumor, and treatment-related effects including radiation necrosis that may not be apparent on anatomical MRI alone by quantifying metabolites [2-5]. However, the utility of MRS critically depends on accurate placement of the spectroscopy volume-of-interest (VOI) based on different clinical contexts. In routine clinical practice, VOI placement is typically performed manually by experienced operators using anatomical images. This requires careful balancing of competing objectives such as maximizing target tissue coverage, avoiding regions including necrosis and cerebrospinal fluid that degrade spectral quality, minimizing partial-volume contamination, and achieving adequate signal-to-noise ratio, which makes VOI placement challenging and time consuming [6-8]. This process also exhibits inter-operator variability, particularly in brain tumors where heterogeneous subregions complicate the placement. It is often the case that multiple spatially distinct VOI placements may be acceptable and satisfy the clinical and technical constraints [9, 10]. Moreover, VOI placement is not governed solely by operator preference, but also by multiple clinically motivated objectives that vary with the specific diagnostic or research question. Technical factors further complicate VOI placement: MR sequence design, shimming performance, coil configuration, and $B_0$ magnetic field strength influence spectral quality, and preferred VOI geometries may therefore differ across systems and institutions. Accommodating this diversity using conventional supervised learning requires retraining separate models for each clinical objective, question, and system configuration, which is impractical.

Prior efforts to automate MRS VOI placements in brain have explored discrete optimization-based search and supervised deep learning approaches [8, 11]. While these methods can produce reasonable placements, they often encode a fixed optimization preference and cannot incorporate context-dependent user instructions. Recent advancements in large language models (LLMs) offer a complementary paradigm [12, 13]. Rather than directly predicting VOI parameters in one shot, an LLM can serve as a high-level controller that interprets user preferences, then selects and calls domain-specific tools [14-16]. In this work, we introduce an agentic LLM workflow for MRS VOI placements in brain tumors that separates the problem into (i) generating diverse, acceptable VOI candidates and (ii) selecting from the candidates using natural language guidance. The workflow is designed as a tool-oriented pipeline in which the LLM coordinates the provided toolsets for segmentation and VOI placement, and integrates quantitative evaluations.

Our novel contributions are as follows: (1) we use an objective function to generate deterministic VOI placements as a supervised training dataset, capturing different human experts' preferences in VOI placement without extensive manual annotation; (2) we propose vision transformer-based and position-conditioned deep learning models to produce diverse candidates of VOIs; (3) we present an agentic LLM workflow for MRS VOI placement that combines segmentation, VOI placement, and LLM-based quantitative metric evaluation; and (4) we show preference-conditioned control, where natural language requests produce targeted VOI solutions for specific clinical objectives.

## 2  Methods

### 2.1  Large Language Model for Agentic Workflow

We use Qwen3-14B [17] as the LLM that mediates the agentic workflow (Fig. 1). The workflow is designed as a tool-oriented pipeline in which the LLM acts as a controller that interprets user preferences, orchestrates tools, and integrates intermediate results to reach a final decision. Four tools are available to the LLM: tumor segmentation, VOI placement, metrics evaluation, and task completion. Given a user instruction specifying VOI placement preferences, the LLM reasons and determines which tool is necessary at each stage, and invokes domain-specific tools through structured tool calls.

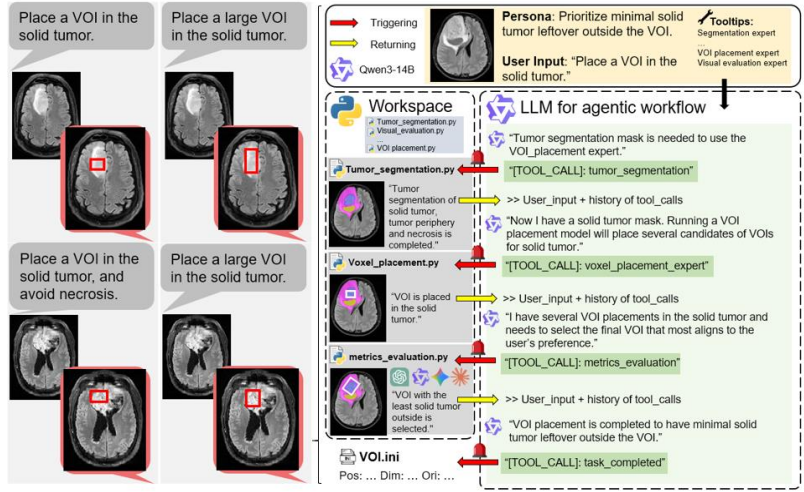

**Fig. 1.** Examples of VOI placement in brain (left) via an agentic workflow (right).

### 2.2 Human Experts' VOI Placement in Brain Tumor

Retrospective manual VOI placements were performed by five MRS experts, three neuroradiologists and two physicists, recruited from five different institutions, with prior MRS experience in clinical practice and research ranging from 15 to 28 years. Imaging data were retrospectively assembled from routine clinical scans of 125 adult glioma patients – low-grade (n=60), high-grade (n=64), and one case with undetermined grade (n=1) – which were acquired at two sites (ANONYMIZED and ANONYMIZED). The 125 cases were randomly partitioned into five blocks of 25 cases, with each block then randomly assigned to two experts. VOIs were positioned on paired $T_1$-weighted and $T_2$-weighted FLAIR images using internally developed software. Consequently, each expert had completed two blocks, resulting in two independent VOI placements for every case. After full data review, 15 studies were excluded for failing the inclusion criteria; 6 showed substantial motion artifacts, 2 had post-operative scans, and 7 had post-contrast acquisitions, which left 110 cases for the final analysis. The study protocol has received ethics board approval, and written informed consent was either obtained from participants included in the study or was waived by the IRB.

### 2.3 Data Augmentation using Discrete Search Objective

Equation (1) defines an objective for VOI placement that prioritizes coverage of specific tumor subregions. Rather than using a homogeneous target, the objective scores a candidate VOI with respect to multiple classes, including solid tumor, tumor periphery, necrosis, and normal brain. For each class, we first compute the fraction of the VOI containing the subregion and then transform this fraction using a Gaussian-shaped function. The base objective is defined as the product of these Gaussian terms across subregion classes. In addition to these tissue fraction terms, we incorporate three geometric modifiers that are multiplicatively combined with the Gaussian objective. A penalty term based on the amount of solid tumor missed outside the VOI favors placements that include sufficient solid tumor to maintain adequate MRS signal-to-noise ratio. Conversely, a term that depends on the VOI volume occupied by solid tumor discourages overly aggressive expansion, preventing the VOI from becoming unnecessarily large. Finally, a distance-to-skull term penalizes VOIs positioned too close to the skull only when distance is less than 5 mm, reflecting degradation in shimming and spectral quality near bone. Thus, the VOIs in brain tumor are evaluated by the fraction of the VOI (*fVOI*) and volume of the VOI (*vVOI*) containing the tumor subregions. Namely, fraction of VOI containing solid tumor, tumor periphery, necrosis or normal brain (*fVOI$_{solid}$*, *fVOI$_{periphery}$*, *fVOI$_{necrosis}$*, *fVOI$_{normal}$*), fraction of solid tumor outside the VOI (*fSolid$_{outside}$*) and volume of the VOI containing necrosis (*vVOI$_{necrosis}$*). The final objective function for *balanced* preference is defined as

$$\prod_c exp\left(-\frac{1}{2}\left(\frac{F_c(\theta)-\mu_c}{\sigma_c}\right)^2\right) \times exp(\mu_V - V) \times exp(\mu_D - D), \quad (1)$$

where $F_c(\theta)$ denotes the geometric overlap between VOI and different tissue compartments parameterized by $\theta$, and ($\mu_c$, $\sigma_c$) specify the desired coverage and tolerance for that class. Parameter values are selected empirically based on distributions from 110 manually placed MRS VOIs: *fVOI$_{solid}$* (54 ± 32%), *fSolid$_{outside}$* (62 ± 26%), *fVOI$_{periphery}$* (32 ± 31%), *fVOI$_{necrosis}$* (4 ± 12%), and *fVOI$_{normal}$* (10 ± 16%). Additionally, V = volume (mL) of the solid tumor inside VOI and D = distance (mm) from skull to VOI.

Two parameter sets are defined to generate distinct training datasets with different *preferences*. The *balanced* preference is designed to reflect the average behavior of human experts who weighted tumor coverage and necrosis avoidance equally, while the *large-VOI* preference was designed to emulate experts who favor larger VOI size. For the *balanced* preference, $\sigma_c$ values are set according to the observed standard deviations, with $\mu_c = 1$ for solid tumor inclusion and $\mu_c = 0$ for all other tissue classes. For the large-VOI preference, $\sigma_c$ values are doubled for all tissue classes except the solid tumor outside term, for which $\sigma_c$ is halved to more strongly penalize missed solid tumor. For both preferences, exponential penalty for VOIs located within 5 mm of the skull is imposed, and only for the *balanced* preference, additional constraint limiting the VOI to contain no more than 15 mL of solid tumor is imposed to prevent generation of a too large VOI with $\mu_V = 15$ and $\mu_D = 5$. 500 cases from the BraTS2021 dataset [17] are randomly sampled and used as the development set.

### 2.4   AI Models in Toolsets

**Tumor Segmentation.** The BraTS 2017 dataset [19-21] is used to train nnUNet [22] to create the tumor segmentation tool. The nnUNet is trained to segment T$_2$w FLAIR images into contrast-enhancing (CE) tumor, non-enhancing (NE) tumor, and necrosis, with standard nnUNet training process with the 5-fold cross validation, which resulted in a Dice score of 0.80. In the cases which have both CE and NE regions, the solid tumor subregion is defined as the CE region, and tumor periphery as NE region, while in the cases which only have NE regions, solid tumor is defined as the NE region.

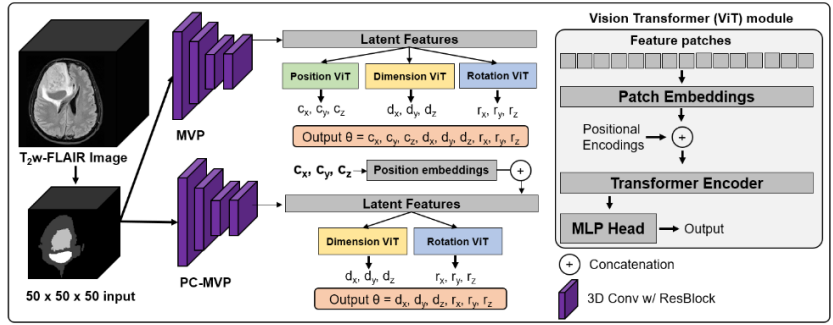

**Fig. 2.** Schematic of MVP and PC-MVP models.

**VOI Placement.** We propose an MRS VOI placement (MVP) model that parameterizes a 3D VOI placement using a 9D vector $\theta$ comprising the VOI center, three side lengths, and Euler-angle rotation (Fig. 2). A network is trained to regress $\theta$ directly from input segmentation $M$, using a weighted $L_2$ loss that separately penalizes errors in position, size, and orientation objective $\lambda \times L(\theta, M)$. Based on initial tuning, we have fixed the loss weights to $\lambda_{pos} = 5$, $\lambda_{dim} = 3$, and $\lambda_{ori} = 2$ for all subsequent experiments. Our vision transformer (ViT) regressor uses 3D convolutional feature extraction followed by independent transformer regression heads to predict VOI position, dimensions, and orientation. Specifically, the tumor segmentation is cropped to a $50 \times 50 \times 50$ volume centered on the segmented tumor and passed through a four-stage 3D residual CNN backbone. The resulting latent representation is routed to three ViT regression heads that output position, dimensions, and orientation, respectively.

We also implement a position-conditioned variant (PC-MVP) to produce a diverse set of multiple acceptable VOIs by sampling different conditioning positions. PC-MVP takes an additional coordinate input representing the intended VOI center; given this position, the model predicts only the corresponding dimensions and orientation for that position. For PC-MVP, position coordinates are sampled at 0.5 mL intervals across the solid tumor. To limit candidate set size during the LLM selection, the total number of sampled coordinates is capped at 50. If the initial 0.5 mL interval produces more than 50 candidates, the sampling interval is adaptively increased so that no more than 50 coordinates are generated. This interval and cap reflect practical constraints of LLM-based decision-making, as presenting an excessive number of candidates VOIs and associated quantitative metrics can reduce LLM effectiveness and increase inference-time memory requirements. The MVP model trained with the *balanced* dataset and the PC-MVP model trained with the *balanced* or large-VOI dataset are all used to generate candidates of VOI placements when the tool is called.

The training and validation datasets for MVP and PC-MVP models consist of 300 and 200 cases, randomly sampled from the BraTS 2021 dataset, with VOIs placed using the specified objective function. For the test set, we have used data from 110 patients with brain tumors that were collected from ANONYMIZED and ANONYMIZED in order to compare the VOI placement by the MVP and PC-MVP models to the human experts. All images are resampled to a standardized spatial resolution of $2 \times 2 \times 2$ mm³ and to a uniform size of 128 × 128 × 128. We have used a batch size of 1, and the Adam optimizer [23] with a 0.0001 learning rate to train for 120,000 iterations.

**LLM-based VOI Selection.** The Qwen3-14B model is given a prompt to select the final VOI placement from the provided candidates based on the tumor subregions metrics. The prompt has specified the characteristics of each MVP and PC-MVP model, and that the MVP model's placement is used as a reference placement. BraTS 2021 cases are used to develop the prompts for the LLM.

## 3    Results and Discussions

### 3.1    Discrete Objective Search for Data Augmentation

Table 1 compares brain tumor VOI placement performance when using an objective function with different preferences on the BraTS2021 development set. Under the *balanced* preference, the objective function achieves solid tumor coverage of $fVOI_{solid}$ = 0.94 ± 0.11 and $fSolid_{outside}$ = 0.80 ± 0.13. When the objective function explicitly prioritizes low $fSolid_{outside}$, $fSolid_{outside}$ becomes as low as 0.60 ± 0.21, which comes at the cost of reduced solid tumor coverage $fVOI_{solid}$ = 0.86 ± 0.15.

**Table 1.** VOI placements of objective function in BraTS2021 development set (n=500).

| Methods | Preference | $fVOI_{solid}$ ↑ | $fSolid_{outside}$ ↓ | $vVOI_{necrosis}$ (mL) ↓ | VOI volume (mL) |
|---|---|---|---|---|---|
| Objective function | Balanced | 0.94 ± 0.11 | 0.80 ± 0.13 | 0.03 ± 0.08 | 16.0 ± 4.1 |
| Objective function | $fSolid_{outside}$ ↓ | 0.86 ± 0.15 | 0.60 ± 0.21 | 0.04 ± 0.09 | 32.0 ± 10.2 |

**Table 2.** VOI placements of experts and discrete search method in the *in vivo* test set (n=110).

| Methods | Preference | $fVOI_{solid}$ ↑ | $fSolid_{outside}$ ↓ | $vVOI_{necrosis}$ (mL) ↓ | VOI volume (mL) |
|---|---|---|---|---|---|
| Human Experts | Balanced | 0.54 ± 0.32 | 0.62 ± 0.26 | 0.59 ± 2.50 | 12.4 ± 6.6 |
| Discrete search | Balanced | 0.74 ± 0.32 | 0.55 ± 0.23 | 0.15 ± 0.21 | 10.6 ± 3.7 |
| Discrete search | $fSolid_{outside}$ ↓ | 0.63 ± 0.20 | 0.33 ± 0.22 | 0.61 ± 1.10 | 19.8 ± 11.2 |

### 3.2    VOI Placements in the *In Vivo* Glioma Cases

Table 2 summarizes VOI placement performance for human experts and discrete search method on the test set (n=110). Average behavior of the five experts' placements represents a *balanced* preference, which achieves moderate solid tumor coverage ($fVOI_{solid}$ = 0.54 ± 0.32 and $fSolid_{outside}$ = 0.62 ± 0.26). The discrete search for *balanced* preference increases $fVOI_{solid}$ while reducing $fSolid_{outside}$ compared to human experts but the average VOI volume becomes 2 mL smaller. When optimized to minimize $fSolid_{outside}$, the method further reduces $fSolid_{outside}$ but at the cost significantly increased VOI volume. These results demonstrate that discrete search can strongly optimize a selected metric, but doing so alters the placement pattern that differs from the human experts.

MVP and PC-MVP models are trained on objective function–based VOI placements derived from the BraTS2021 dataset, reflecting different priorities of *balanced* and low $fSolid_{outside}$. Under the *balanced* preference, both MVP and PC-MVP achieve better performance than human experts and previously reported automated approaches with CNN-based model [9] and discrete search [8] in terms of $fVOI_{solid}$ and $fSolid_{outside}$ (Table 3). When trained with low $fSolid_{outside}$ prioritized datasets, both MVP and PC-MVP substantially reduce solid tumor exclusion compared to *balanced* training and prior approaches, at the cost of increased VOI volume.

**Table 3.** VOI placements of human experts and DL methods in the *in vivo* test set (n=110).

| Methods | Preference | $fVOI_{solid}$ ↑ | $fSolid_{outside}$ ↓ | $vVOI_{necrosis}$ (mL) ↓ | VOI volume (mL) |
|---|---|---|---|---|---|
| Lee et al. [9] | Balanced | 0.56 ± 0.30 | 0.61 ± 0.21 | 0.80 ± 1.92 | 11.6 ± 3.4 |
| Bolan et al. [8] | Balanced | 0.48 ± 0.29 | 0.62 ± 0.25 | 0.72 ± 1.82 | 11.9 ± 2.1 |
| MVP | Balanced | 0.60 ± 0.29 | 0.59 ± 0.25 | 0.57 ± 1.45 | 10.5 ± 3.1 |
| PC-MVP | Balanced | 0.51 ± 0.29 | 0.61 ± 0.25 | 0.53 ± 1.39 | 11.5 ± 2.9 |
| MVP | $fSolid_{outside}$ ↓ | 0.45 ± 0.26 | 0.37 ± 0.25 | 1.5 ± 3.8 | 24.9 ± 7.2 |
| PC-MVP | $fSolid_{outside}$ ↓ | 0.38 ± 0.24 | 0.40 ± 0.27 | 1.1 ± 2.9 | 28.5 ± 6.8 |
| Agentic workflow | $vVOI_{necrosis}$ ↓ | 0.64 ± 0.25 | 0.55 ± 0.22 | **0.35 ± 0.59** | 12.4 ± 3.7 |
| Agentic workflow | $fVOI_{solid}$ ↑ | **0.66 ± 0.23** | 0.55 ± 0.21 | 0.45 ± 0.76 | 11.8 ± 3.2 |
| Agentic workflow | $fSolid_{outside}$ ↓ | 0.63 ± 0.25 | **0.52 ± 0.22** | 0.53 ± 1.10 | 14.1 ± 6.1 |

The example in Figure 3 shows how PC-MVP model and preference-specific training datasets expand VOI placement beyond a single deterministic solution. The baseline MVP model trained on a *balanced* dataset (red VOI) produces one deterministic placement. In contrast, PC-MVP model is conditioned on all sampled positions, where one of them is shown as a green dot, resulting in multiple distinct placements. When trained on the same *balanced* dataset (blue VOI), PC-MVP generates a placement similar in scale and location to MVP, with minor shifts in orientation and dimension. When trained on a dataset curated toward low $fSolid_{outside}$ (yellow VOI), PC-MVP expands coverage toward undersampled tumor regions, reducing missed solid tumor at the cost of getting closer to the skull. This ability to generate multiple valid VOIs within the solid tumor is essential for the downstream agentic LLM workflow, which selects among candidates based on explicit clinical preferences.

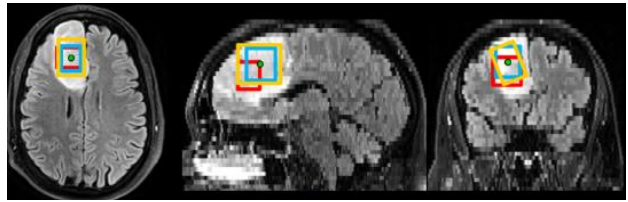

**Fig. 3.** Different VOIs placed by MVP model trained with balanced dataset (red), and PC-MVP model trained with balanced (blue) and low $fSolid_{outside}$ dataset (yellow) for a case in the test set. Blue and yellow VOIs share the same position coordinate (green dot).

The agentic workflow demonstrates preference-dependent behavior. Table 3 shows that the agentic workflow achieves the lowest necrotic tissue inclusion (0.35 ± 0.59 mL) among all methods with the low $vVOI_{necrosis}$ preference, the highest solid tumor coverage (0.66 ± 0.23) with the high $fVOI_{solid}$ preference, and the lowest $fSolid_{outside}$ of 0.52

± 0.22 with the low $fSolid_{outside}$ preference. Importantly, these metric improvements are achieved without substantially altering the overall placement pattern relative to human experts. Across preferences, the VOI volume remains within a similar range, approximately 11.8–14.1 mL, comparable to expert placements of around 12 mL, rather than exhibiting significant change in VOI volume observed in discrete search method.

Figure 4 demonstrates how the proposed agentic workflow reproduces expert-like VOI placement behavior by conditioning different preferences. On the left, two human experts exhibit distinct preferences. Expert 1 constrained the maximum VOI volume, resulting in a smaller VOI that targets the tumor core, but leaves a large fraction of solid tumor outside the VOI, whereas Expert 2 favors a larger VOI for this large tumor, increasing coverage at the cost of VOI size. On the right, the agentic workflow produces analogous placements when guided by corresponding preferences. A preference emphasizing increased $fVOI_{solid}$ yields a compact VOI comparable to Expert 1, while a preference preferring lower $fSolid_{outside}$ generates a larger VOI that mirrors Expert 2's strategy. These results illustrate that the agentic workflow can modulate VOI size to align its outputs with different expert preferences.

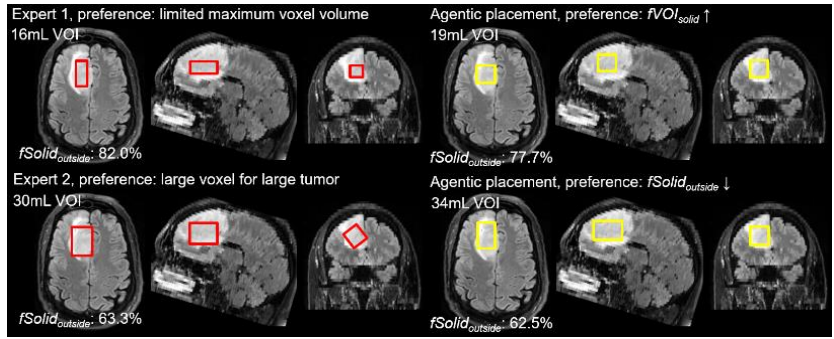

**Fig. 4.** Example VOIs placed by human experts and agentic workflow for a case in the test set.

## 4   Conclusion

We have presented an agentic LLM workflow for MRS VOI placement in brain tumors, demonstrating that natural language preferences can be translated into different VOI configurations by selecting among diverse candidate placements. Our proposed agentic workflow is designed to accommodate the variability by multiple candidate generation and LLM's selection without retraining separate models for each clinical question while still maintaining the human experts' placement pattern. Future work will evaluate VOI placements across a broader range of clinical contexts and technical settings and, as a longer-term direction, extend the framework toward a foundational model capable of comprehensive VOI placement in normal brain anatomy.

**Acknowledgments.** Authors would like to thank Sarah Bedell, Noam Harel, and Henry Braun for help with de-identification of images. This work was supported by NIH grants U01CA269110, R01EB034231, and P41 EB027061. FB acknowledges support from Investissements d'avenir [grant numbers ANR-10-IAIHU-06 and ANR-11-INBS-0006] and from Agence Nationale de la Recherche [grant number ANR-20-CE17-0002-01].

**Disclosure of Interests.** The authors have no competing interests to declare.